\documentclass[letterpaper, 10 pt, conference]{ieeeconf} 

\IEEEoverridecommandlockouts                             
                                                          
\usepackage{multicol}
\usepackage[bookmarks=true]{hyperref}
\usepackage{booktabs}
\usepackage{tabularx}
\usepackage{tabulary}
\usepackage{array}
\usepackage{multirow}
\usepackage{url}            
\usepackage{graphicx} 
\usepackage[table,dvipsnames]{xcolor}
\usepackage{amssymb}
\usepackage{amsmath}
\usepackage{cuted}
\usepackage{gensymb}
\usepackage{multirow}
\usepackage{caption}
\usepackage{pifont}

\usepackage[noadjust]{cite}

\usepackage[most]{tcolorbox}
\definecolor{mygray}{gray}{.9}

\hypersetup{
    colorlinks=true,
    linkcolor=blue,
    filecolor=magenta,      
    urlcolor=cyan,
}

\newcommand{\website}{https://github.com/Yunheng-Wang/DreamNav}

\newcolumntype{P}[1]{>{\centering\arraybackslash}p{#1}}

\title{\LARGE \bf
DreamNav: A Trajectory-Based Imaginative Framework for Zero-Shot Vision-and-Language Navigation}

\author{%
Yunheng Wang$^{1,*}$, Yuetong Fang$^{1,*}$, Taowen Wang$^{1,*}$, Yixiao Feng$^{1,*}$, Yawen Tan$^{2}$,\\
Shuning Zhang$^{1}$, Peiran Liu$^{1}$, Yiding Ji$^{1}$, Renjing Xu$^{1,\dagger}$%
\thanks{$^{1}$ The Hong Kong University of Science and Technology (Guangzhou).}%
\thanks{$^{2}$ Zhejiang Normal University.}%
\thanks{$^{*}$ Equal contribution. \texttt{yunhengwang1214@gmail.com}}%
\thanks{$^{\dagger}$ Corresponding author. \texttt{renjingxu@hkust-gz.edu.cn}}%
}

\begin{document}

\maketitle
\thispagestyle{empty}
\pagestyle{empty}

\vspace{15pt}  


\begin{abstract}

Vision-and-Language Navigation in Continuous Environments~(VLN-CE), which links language instructions to perception and control in the real world, is a core capability of embodied robots. Recently, large-scale pretrained foundation models have been leveraged as shared priors for perception, reasoning, and action, enabling zero-shot VLN without task-specific training. However, existing zero-shot VLN methods depend on costly perception and passive scene understanding, collapsing control to point-level choices. As a result, they are expensive to deploy, misaligned in action semantics, and short-sighted in planning. To address these issues, we present DreamNav that focuses on the following three aspects:~(1) for reducing sensory cost, our EgoView Corrector aligns viewpoints and stabilizes egocentric perception;~(2) instead of point-level actions, our Trajectory Predictor favors global trajectory-level planning to better align with instruction semantics; and~(3) to enable anticipatory and long-horizon planning, we propose an Imagination Predictor to endow the agent with proactive thinking capability. On VLN-CE and real-world tests, DreamNav sets a new zero-shot state-of-the-art~(SOTA), outperforming the strongest egocentric baseline with extra information by up to 7.49\% and 18.15\% in terms of SR and SPL metrics. To our knowledge, this is the first zero-shot VLN method to unify trajectory-level planning and active imagination while using only egocentric inputs.

\end{abstract}

\section{Introduction}

 \begin{figure}[t]
   \centering
   \includegraphics[width=1\linewidth]{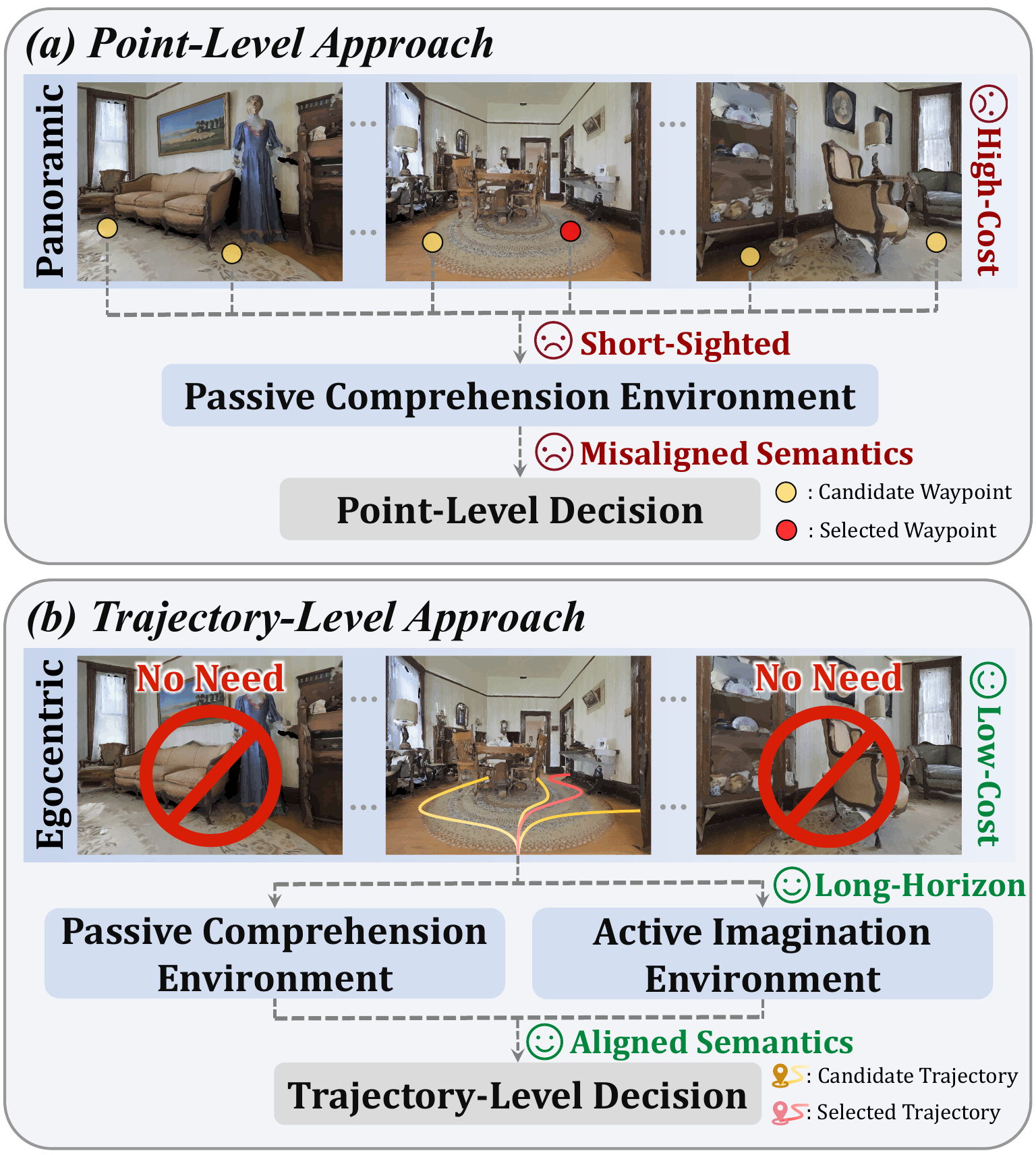}
   \vspace{-15pt}
      \caption{\footnotesize Comparison between the traditional point-level approach and our trajectory-level approach. (a) Point-level approach passively interprets panoramic inputs before selecting candidate points, leading to high acquisition cost, short-sightedness, and misaligned semantics. In contrast, (b) Trajectory-level approach leverages egocentric inputs to actively imagine future scenarios and select candidate trajectories, achieving low acquisition cost, long-horizon, and aligned semantics.}
    \label{fig:introduction}
    \vspace{-15pt}
\end{figure}




Vision-and-Language Navigation in Continuous Environments (VLN-CE)~\cite{vlnce} refers to a task where an agent follows natural language instructions to perceive its surroundings and navigate through a continuous visual space to reach a target location. Unlike discrete VLN tasks~\cite{vln}, VLN-CE can work beyond predefined nodes or maps with direct interaction, enabling agents to directly couple environmental perception with language understanding. This makes navigation more flexible and more relevant to real-world scenarios. Therefore, VLN-CE represents a crucial step toward building embodied agents that can operate reliably in unstructured and dynamic real-world environments. This further drives VLN research toward its central goal: grounding natural language in embodied action.

Early VLN progressed through modular, training-heavy pipelines that focus on four perspectives: representation, strategy, data augmentation, and exploration~\cite{vln_survey}. Recent works have begun to leverage large pretrained multimodal models for end-to-end learning, e.g., fine-tuning vision–language models~\cite{vln_r1} or mapping video to actions~\cite{navid,uni_navid}. This trend reduces hand engineering and streamlines training. Similarly, the powerful foundation models can also serve as the cognitive core for VLN systems. A clear benefit is that foundation models have already been pre-trained on Internet-scale data and can encode broad world knowledge and structural priors. This enhanced cognitive capability enables VLNs to perform zero-shot reasoning and in-context learning across tasks without fine-tuning~\cite{mp5}. The demonstrated success in discrete zero-shot VLN~\cite{navgpt,mapgpt} naturally motivates extending the strengths of foundation models to zero-shot VLN in continuous environments. 

Action strategies in zero-shot VLN within continuous environments are divided into point-level and trajectory-level categories. We provide an intuitive comparison between point-level and trajectory-level VLN systems as shown in Fig.~\hyperref[fig:introduction]{1}. Most of the prior approaches currently heavily rely on point-level actions~\cite{opennav,smartway,instructnav}: the agent samples multiple candidate points from high-level panoramic observations, passively comprehends the aggregated local surroundings, and then makes point-level decisions. While this offers a bridge from discrete to continuous environments, the paradigm has inherent limitations: ~\ding{182} High-cost: Panoramic observation necessitates either multi-sensor configurations, entailing substantial financial cost, or single-camera in-place rotations, imposing pronounced temporal cost;~\ding{183} Short-sightedness: passive comprehension restricts decisions to current and past views, lacking foresight for future reasoning or long-horizon planning;~\ding{184} Misaligned semantics: point-level decisions are weakly aligned with instruction-level global semantics, making the policy prone to locally optimal decision-making~\cite{View_Invariant_Learning}.

\begin{figure*}[t!]
\centering
\includegraphics[width=1\linewidth]{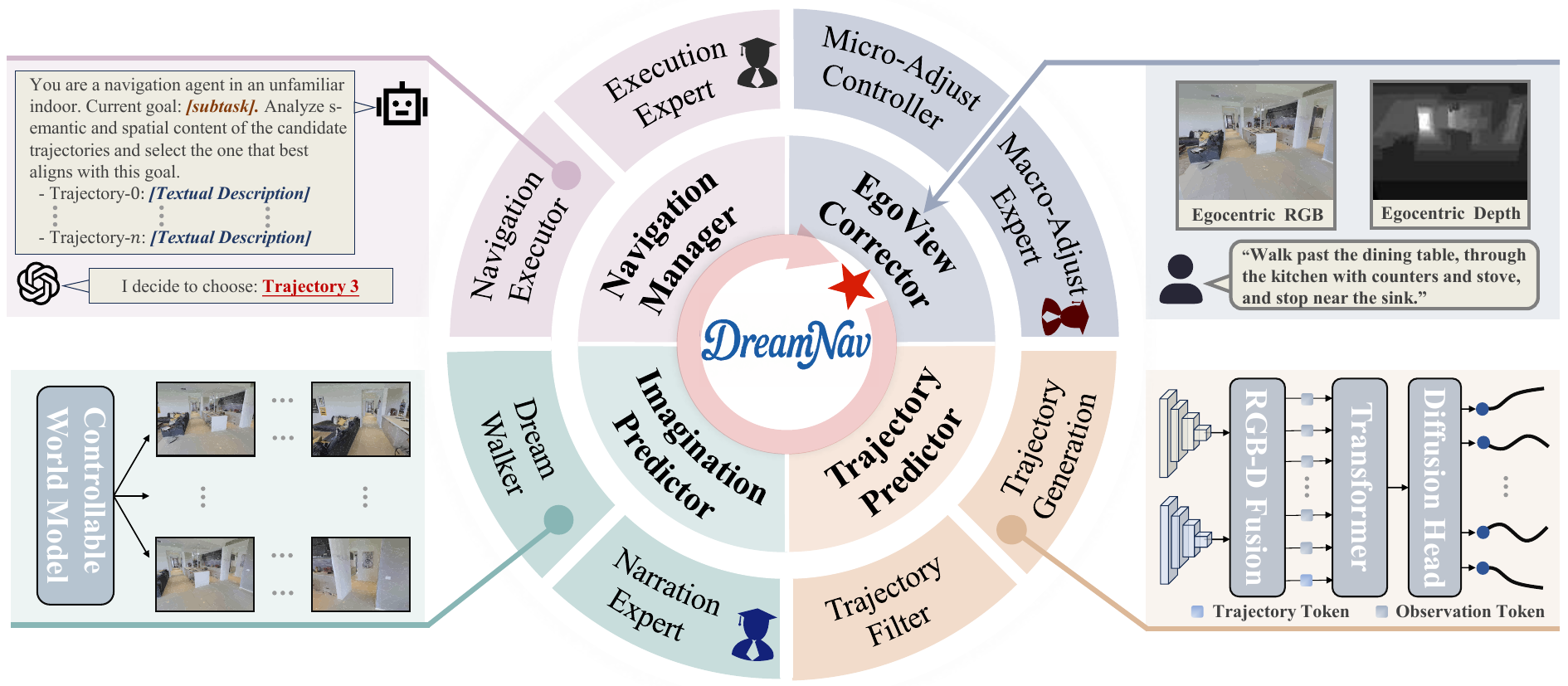}
   \vspace{-15pt}
    \caption{\footnotesize\textbf{Pipeline of DreamNav.} Our approach is structured into four hierarchical modules, each instantiated as a pair of mutually coupled submodules. It takes egocentric RGB-D observations and a natural language instruction as input. The EgoView Corrector first aligns the agent’s viewpoint with the instruction-indicated orientation. The Trajectory Predictor then generates candidate trajectories, and the Imagination Predictor evaluates their long-horizon outcomes. Finally, the Navigation Manager integrates these predictions to select and execute the optimal trajectory while monitoring task progress.}
    \label{fig:method_1}
    \vspace{-15pt}
\end{figure*}

Human navigation adapts to unfamiliar environments by constructing coherent trajectories and actively imagining possible futures. Yet, it remains an open question whether incorporating these human-like capabilities can advance VLN-CE systems. Motivated by this, we present DreamNav that unifies trajectory-level planning and active imagination in zero-shot VLN systems as illustrated in Fig.~\ref{fig:method_1}. DreamNav stabilizes low‑cost egocentric perception with an EgoView Corrector that uses a hierarchical scheme to progressively align observations with panoramic sensing. With this efficient perceptual scheme, 
we equip agents with an Imagination Predictor that employs controllable visual generation and video understanding models to transform passive comprehension into active imagination for long-horizon reasoning. A Trajectory Predictor further integrates diffusion‑based trajectory generation with geometric filtering to maintain instruction‑level coherence, and the Navigation Manager then unifies these predictions into optimal trajectories executed with continuous progress monitoring.
We summarize our contributions as follows:

\begin{itemize}
    
    \item We alleviate the \textbf{high-cost perception} challenge by designing an EgoView Corrector in DreamNav that operates solely on low-cost egocentric inputs.
    
    \item We mitigate the \textbf{short-sightedness} challenge through an Imagination Predictor that activates prospective rollouts, enabling long-range reasoning in zero-shot VLNs.
    
    
    \item We resolve the \textbf{misaligned semantic} challenge by introducing a Trajectory Predictor that generates trajectory-level action policies aligned with semantics to ensure globally coherent navigation strategies. 
    
  
\end{itemize}

\section{Related Works}
\label{sec:related_works}

\subsection{Perception in Zero-Shot VLN Agents}
Perception serves as the primary channel through which zero-shot VLN agents understand and interact with their environments. Most existing zero-shot VLN approaches~\cite{navgpt,mc_gpt,mapgpt,instructnav,smartway,opennav} rely on panoramic perception, which provides comprehensive global scene context and has driven substantial progress. For instance, NavGPT~\cite{navgpt} and OpenNav~\cite{opennav} exploit the capabilities of GPT-4 with panoramic visual inputs to explore the feasibility of zero-shot VLN in discretized and continuous environments~\cite{vlnce}, respectively. However, panoramic inputs entail high costs that hinder their practicality in zero-shot VLN, incurring substantial acquisition overhead from multi-sensor configurations or exhaustive in-place rotations, and introducing significant visual redundancy that imposes heavy computational burdens on downstream reasoning. An appealing alternative is to rely on egocentric visual inputs, yet their limited perceptual coverage and susceptibility to orientation errors have hindered progress, resulting in sparse research on zero-shot VLN with egocentric perception. CA-Nav~\cite{ca_nav} is the only work exploring egocentric inputs; however, its reliance on camera poses limits real-world applicability due to their inaccessibility and low accuracy in indoor settings. To address the high perception cost, our DreamNav designs an EgoView Corrector, which mitigates viewpoint misalignment to bridge the gap to panoramic observation, thereby enabling low-cost zero-shot VLN navigation.

\subsection{Action Strategies for Zero-Shot VLN}


Existing zero-shot VLN in continuous environment methods~\cite{opennav,smartway} commonly adopt the point-level action strategy, which requires a pre-trained waypoint predictor~\cite{bridging} to discretize the continuous space by directly inferring the next navigable point from visual observations. While this has led to notable progress, restricting actions to straight-line waypoint moves narrows the navigable space and misaligned instruction semantics, thus often leading to locally optimal decisions. To address this issue, some approaches~\cite{instructnav,ca_nav} attempt to predict candidate decision points through multi-sourced value maps for long-horizon, instruction-aligned decisions. Nevertheless, a new challenge arises from the non-strict geometric constraints and their strong reliance on semantic segmentation and object recognition, which in turn causes imprecise waypoint proposals, reduced navigation accuracy, and increased collision risk. In this work, we aim to develop zero-shot VLN systems that are long-horizon, instruction-aligned semantics, and collision-aware. Unlike prior studies, our proposed DreamNav introduces a Trajectory Predictor that performs trajectory-level action planning to seek globally coherent decision-making.


\subsection{Imagination for VLN}
While zero-shot VLN systems based on passive scene understanding have made progress, their reliance on restricted inputs and passive reasoning creates inherent constraints that hinder performance.  In response, several studies have explored imagination-based strategies in VLN, including predicting future visual observations with learned world models~\cite{pathdreamer, dreamwalker, nwm}, inferring occupancy and semantic layouts from partial maps~\cite{imagine_before_go, peanut}, and forecasting future semantic features~\cite{unitedvln, lookahead}. However, these approaches are mostly designed as auxiliary components or are tightly coupled with supervised paradigms, making them less suitable for zero-shot VLN. This is because feeding generated visual data into large models incurs unpredictable API costs, and such models are ill-suited to interpret high-level feature representations or localized maps. Thus, it is crucial to design an imagination module tailored for zero-shot VLN.

To overcome the short-sightedness inherent in passive comprehension, we introduce an Imagination Predictor in DreamNav, which encodes imaginative priors into structured textual descriptions aligned with foundation models, thereby endowing agents with long-horizon foresight and planning.

\section{METHOD}
In this section, we first define the task (Sec.\ref{sec:problem_formulation}) and then describe DreamNav, a closed-loop anthropomorphic indoor navigation system. As illustrated in Fig.~\ref{fig:method_1}, the system takes egocentric RGB-D observations and natural-language instructions as inputs, and comprises four key components: EgoView Corrector (Sec.~\ref{sec:egoview_corrector}) mitigates orientation errors by realigning egocentric observations for stable perception; Trajectory Predictor (Sec.~\ref{sec:trajectory_predictor}) for planning multiple feasible and navigability-aware route strategies; Imagination Predictor (Sec.~\ref{sec:imaginative_predictor}) enriches candidate trajectories with long-horizon anticipations, extending the agent’s perceptual horizon to proactively infer what lies ahead. Navigation Manager (Sec.~\ref{sec:navigation_manager}) selects the most promising trajectory for execution and reviews progress to ensure task completion.

\subsection{Problem Formulation}\label{sec:problem_formulation}

The zero‑shot VLN problem tasks an agent with navigating from a start position to a designated target solely under the guidance of foundation models, following a natural language instruction $L$, without any task‑specific fine‑tuning. At each step $t$, the agent perceives a visual input $I_t$ and predicts an action $a_{t+1} \in A$, leading to the subsequent state with observation $I_{t+1}$. In this work, the perceptual space $I$ is strictly limited to monocular egocentric RGB‑D streams, while the action space is abstracted to trajectories, each represented as a sequence of high‑level movements extending beyond single‑step actions.


\subsection{EgoView Corrector}\label{sec:egoview_corrector}
\begin{figure}[t]
   \centering
   \includegraphics[width=1\linewidth]{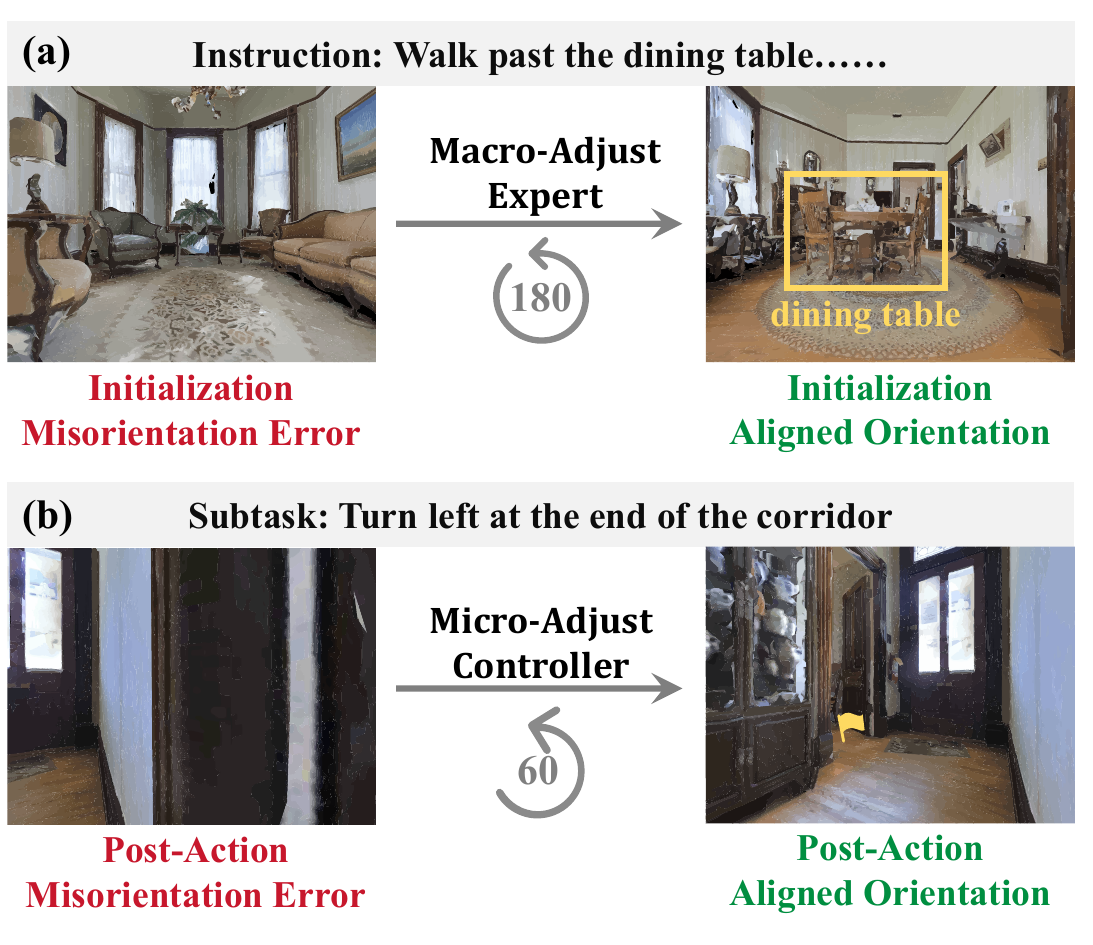}
   \vspace{-15pt}
      \caption{\footnotesize Examples of two types of viewpoint errors encountered by an egocentric agent. (a) At initialization, the agent fails to perceive the instruction-indicated landmark (Initialization Misorientation Error) and is reoriented by the Macro-Adjust Expert to reveal the intended landmark. (b) After action execution, the agent’s view becomes occluded (Post-Action Misorientation Error) and is corrected by the Micro-Adjust Controller via fine adjustment to restore proper orientation.}
    \vspace{-15pt}
    \label{fig:method_2}
    
\end{figure}

Unlike panoramic paradigms, egocentric agents are susceptible to two characteristic viewpoint errors as shown in Fig.~\ref{fig:method_2}. The first error, called the initialization misorientation error, arises at initialization, when the agent’s starting orientation is significantly misaligned with the direction specified by the instruction. The second error, namely the post-action misorientation error, occurs after action execution, when the updated heading leaves the agent unfavorably oriented for planning the next subtask. Both errors have the potential to exclude the target from view, hindering the navigation process. To mitigate these issues, we introduce a two‑stage hierarchical scheme composed of a Macro‑Adjust Expert and a Micro‑Adjust Controller.


\textbf{Macro-Adjust Expert.} Egocentric agents are highly sensitive to the initialization stage itself, where even a modest heading misalignment can compound across steps and ultimately result in task failure. We designed a Macro‑Adjust Expert based on~\cite{achiam2023gpt} to address this problem. Given the raw instruction $L$ and the egocentric observation $I_t^{rgb}$, the expert judges whether the current orientation is already consistent with the instruction or whether a corrective rotation is required, typically 90° clockwise. Its decision emerges from jointly reasoning over the global scene layout and navigable affordances, the presence of instruction‑relevant cues in view, and the directional coherence of the intended subtask. As shown in Fig.~\hyperref[fig:method_2]{3 (a)}, when the initial observation lacks the landmark “dining table,” our Macro‑Adjust Expert correctly identifies this as an Initialization Misorientation Error and performs two successive rotations to realign the agent with the correct orientation.


\textbf{Micro-Adjust Controller.} Egocentric agents executing long-horizon strategies are prone to drift in their terminal heading, manifesting as the Post-Action Misorientation Error. For this problem, we design a Micro-Adjust Controller that performs fine‑grained orientation corrections. Specifically, the controller leverages CLIP‑prompted FastSAM~\cite{fast_segment_anything} to reliably detect the onset of this error and trigger appropriate adjustments.
For instance, given the egocentric $I_t^{rgb}$ and prompt $P_{walk}=$\emph{``walkable indoor area: floor, hallway, corridor, aisle, ramp, stairs''}, FastSAM returns a binary walkable-area mask $\tilde{M}_t$ over the image lattice $\Omega$. The controller then gates reorientation by comparing the normalized mask occupancy to a fixed threshold $\theta \in (0,1)$:
\begin{align}
d_t = 
\begin{cases} 0, & \dfrac{|\widetilde{M}_t|}{|\Omega|} > \theta, \\[6pt] 1, & \dfrac{|\widetilde{M}_t|}{|\Omega|} \le \theta. 
\end{cases} 
\end{align}
We treat $d_t$ as a binary gate that triggers the Post-Action Misorientation error exclusively when $d_t = 1$. Once activated, the current image is partitioned into left and right half‑planes, $\Omega^L$ and $\Omega^R$, and their normalized walkability scores are subsequently compared:
\begin{align}
u_t = \text{sign}\!\left(\frac{|\tilde{M}_t \cap \Omega^L|}{|\Omega^L|} - \frac{|\tilde{M}_t \cap \Omega^R|}{|\Omega^R|}\right)
\end{align}
If $u_t \geq 0$, the agent executes a $30^\circ$ left turn; if $u_t < 0$, a $30^\circ$ right turn. After each adjustment,  both the segmentation mask and gating variable are recomputed, and the routine terminates as soon as $d_t = 0$ or a maximum of two micro‑adjustments has been reached. To mitigate oscillations near the decision boundary, all corrective rotations within a cycle are constrained to follow the same turning direction as the initial adjustment. As illustrated in Fig.~\hyperref[fig:method_2]{3 (b)}, when the agent’s post‑action view is dominated by a wall, a Post‑Action Misorientation Error arises; through two corrective turns, the Micro‑Adjust Controller restores a Post‑Action Aligned Orientation.


\subsection{Trajectory Predictor}\label{sec:trajectory_predictor}
The Trajectory Predictor generates navigation paths that are semantically aligned, traversable, and spatially diverse. It comprises a Trajectory Generator that models the multi-modal distribution of feasible paths and produces diverse candidate trajectories, and a Trajectory Filter that extracts a subset maximizing geometric dissimilarity.

\textbf{Trajectory Generator.} Diffusion‑policy frameworks~\cite{diffusion_policy} have recently demonstrated strong performance in both robotic manipulation~\cite{diffusionvla} and object‑goal navigation~\cite{nomad,ldp}.
Building on these advances, we present the first exploration of their application to zero‑shot VLN, aiming to enable more effective navigation planning.
We employ a Generative Diffusion Policy Model~\cite{navdp}, shown in the light-orange region at the lower-right of Fig.~\ref{fig:method_1}, which takes egocentric RGB-D observations as input, encoding RGB via a pretrained DepthAnything ViT~\cite{depth_anything} ($f_{\text{DEP-ViT}}$) and depth via the NavDP-specific ViT~\cite{navdp} ($f_{\text{RGB-ViT}}$), before fusing the resulting tokens into a unified representation through a transformer decoder ($G_{\text{dec}}$):
\begin{align}
v_t = G_{\text{dec}}\big(f_{\text{RGB-ViT}}(I_t^{rgb}) \, || \, f_{\text{DEP-ViT}}(I_t^{depth})\big)
\end{align}
The fused representations $v_t$ are processed by a two-layer transformer encoder ($\text{G}_{enc}$) to capture spatial relationships and contextual dependencies, forming the conditional context for trajectory generation:
\begin{align}
k_t = \text{G}_{enc}(z_t)
\end{align}
Finally, the diffusion policy head employs a conditional U-Net~\cite{u_net} ($\epsilon_\theta$) with a DDPM scheduler~\cite{denoising} to generate 24 future waypoints 
$\tau = \{\Delta x_t, \Delta y_t, \Delta \text{yaw}_t \}_{t=1}^{24}$ 
by iteratively denoising a perturbed sample $P_t$ drawn from Gaussian noise until obtaining the final noise-free sample $P_0$:
\begin{align}
P_{t-1} = \alpha \Big(P_t - \gamma \, \epsilon_\theta(k_t, P_t, t) + \mathcal{N}(0, \sigma^2, 1)\Big)
\end{align}
where $\alpha, \gamma, \sigma$ are noise schedule of the function.

\textbf{Trajectory Filter.} While the diffusion policy can generate numerous candidates at once, only a small subset corresponds to genuinely distinct and traversable branches, leading to redundancy and limited diversity. Forwarding all candidates downstream substantially increases computational and API token costs, yet contributes only marginal benefits to planning.
To address this, we identify a compact set of candidate trajectories that maximally preserves diversity, ideally assigning one representative trajectory to each traversable branch. This is achieved via the farthest‑first traversal algorithm~\cite{clustering_max_min}, a greedy max-min selection strategy. Specifically, we first quantify pairwise dissimilarity using the mean per-step Euclidean distance:
\begin{align}
d(\tau^i,\tau^j) = \frac{1}{T} \sum_{t=1}^{T} \left\| 
\begin{bmatrix}
\Delta x_t^i - \Delta x_t^j \\
\Delta y_t^i - \Delta y_t^j
\end{bmatrix}
\right\|_2
\end{align}
Given the precomputed distance matrix $D$, we then apply the greedy selection rule:
\begin{align}
s^* = \arg\max_{i \notin S} \min_{j \in S} D_{ij}
\end{align}
We use the candidate trajectory number (CTN) as a hyperparameter that balances diversity preservation against computational efficiency. The iteration proceeds until $|S| = \text{CTN}$.

\subsection{Imagination Predictor}\label{sec:imaginative_predictor}
Designing an imagination module for zero-shot VLN is essential to endow agents with the ability to anticipate future scenarios and enhance the reliability of navigation decisions. However, constructing such a module necessitates consideration of two key factors: (1) API Cost: imagination outputs should minimize overhead in downstream foundation‑model decision‑making; (2) 
Interpretability: imagination outputs should avoid representations that foundation models cannot easily parse or leverage effectively. Therefore, we propose a coupled scheme of Dream Walker and Narration Expert, reformulating imagination entirely in text while eliminating less practical modalities such as images. 


\textbf{Dream Walker.} Imagination should be diverse and unconstrained, motivating us to depart from conventional paradigms that train models in specific indoor environments before deployment in the same domain. As shown in the light-green region at the bottom-left of Fig.~\ref{fig:method_1}, we adopt a controllable world model~\cite{svc} trained on large‑scale, diverse data rather than limited scene data from a specific navigation task.  Given an RGB observation and a trajectory, the model functions as an egocentric simulator, producing coherent visual rollouts conditioned on the trajectory and thereby transforming a static image into an explorable 3D world. Each trajectory element $\tau$ is treated as an atomic action $a_i$ mapped to a relative camera pose $c_i$, enabling $\tau$ to be deterministically translated into the pose sequence $C = {c_0, \ldots, c_{23}}$. We further introduce a hyperparameter called Imagination Rollout Length (IRL), which specifies the horizon of imagination and balances long‑range foresight against accumulated uncertainty. Conditioning the model on each pose in the truncated sequence $C_{0:\mathrm{IRL}}$ ensures that the $i$‑th synthesized frame is aligned with the intended egocentric motion, which can be formally expressed as:
\begin{equation}
W:\,(I_t^{rgb},\, C_{0:\mathrm{IRL}}) \;\;\to\;\; V^{(\mathrm{IRL})} = (V_1^{rgb}, \ldots, V_{\mathrm{IRL}}^{rgb})
\end{equation}
where $V^{(\mathrm{IRL})}$ represents the imagined frame sequence produced by the model.

\textbf{Narration Expert.} The raw pixel‑level rollouts generated by Dream Walker yield fine‑grained detail but incur high computational cost and introduce noise and redundancy.  We argue that effective navigation relies not on exhaustive scene reconstruction, but on identifying task‑relevant semantics. Accordingly, the Narration Expert abstracts dense visual sequences into concise semantic narratives, thereby enhancing efficiency and interpretability. To achieve this, we design targeted prompts composed of guiding questions, including: (1)~What is the agent’s walking direction with respect to the initial viewpoint? (2)~Is the agent approaching a particular object or structural region? (3)~Which reference objects or landmarks does the agent encounter along the path? (4)~How can the overall structural layout of the environment be characterized? (5)~What semantic intent underlies the trajectory? These guidelines explicitly steer the model to focus on task‑critical aspects of the trajectory.



\subsection{Navigation Manager}\label{sec:navigation_manager}
\textbf{Navigator.} The Navigator identifies the most suitable trajectory from multiple imagined candidates to fulfill the current subtask, as shown in the light-pink region at the top-left of Fig.~\ref{fig:method_1}. It integrates the trajectory description generated by the Imagination Predictor with the subtask specification and performs comparative semantic analysis to evaluate their relevance. Each description serves as a predictive signal, allowing the Navigator to assess alignment with the subtask objective and prioritize the candidate that best facilitates task completion. Finally, structured reasoning is generated for all candidates, and the trajectory ID most consistent with the current goal is selected.

\textbf{Execution Expert.} Executing a single navigation subtask typically requires multiple movement steps, which blurs the correspondence between fine-grained actions and higher-level progress. For instance, while the overall navigation trajectory is decomposed into ordered subtasks, the default assumption in VLN is that each movement step completes exactly one subtask. In practice, however, the VLN may advance none, one, or several subtasks at one single step, leading to execution errors that accumulate over the trajectory. To resolve this ambiguity, we extend prior progress‑estimation approaches by introducing an Execution Expert based on~\cite{achiam2023gpt}, which applies a structured chain of reasoning over the recent execution context. Given the previous egocentric view annotated with the actual trajectory aligned to the same perspective, together with the current egocentric view and two ordered subtasks, the expert processes these inputs in sequence. It first contrasts past and current observations to detect instruction‑relevant landmarks, then infers motion from the trajectory annotation, and finally aligns these visual cues with subtask semantics. Through this reasoning process, the expert enforces rigorous subtask execution: maintaining strict sequential fidelity, while accurately discerning task progression by recognizing whether each trajectory step advances zero, one, or multiple subtasks, thereby minimizing perception–execution misalignment.

\section{EXPERIMENTS}
\label{sec:exp}

\begin{figure*}[t!]
\centering
\includegraphics[width=1\linewidth]{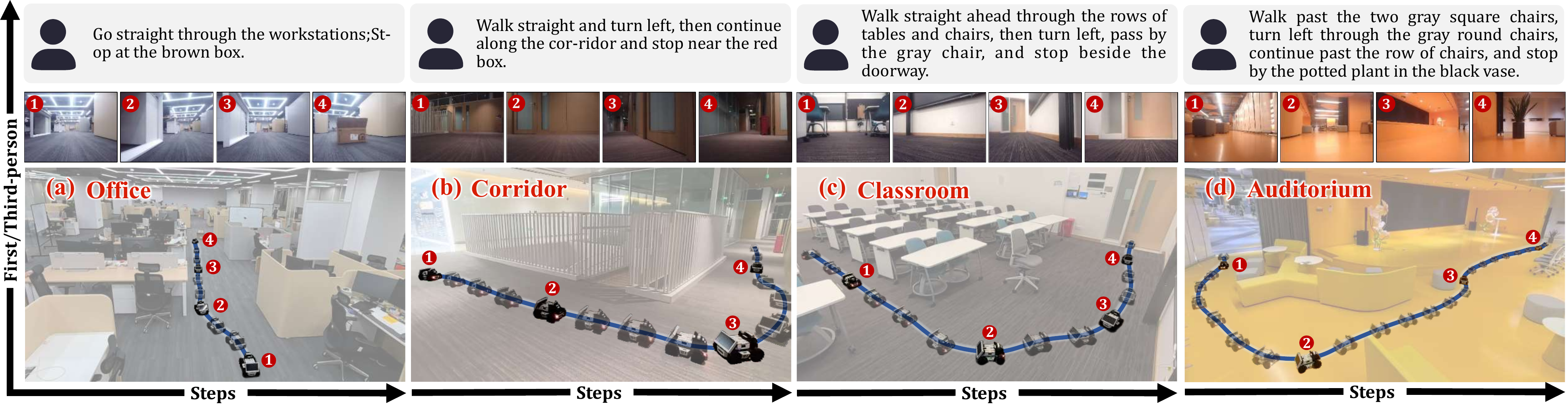}
   \vspace{-15pt}
    \caption{\footnotesize Representative DreamNav navigation examples in real‑world settings: (a) Office, (b) Corridor, (c) Classroom, and (d) Auditorium. For each scene, the top row shows the navigation instruction, the middle row presents the egocentric observations used for execution, and the bottom row depicts the corresponding exocentric (third‑person) trajectory. The middle‑row views correspond directly to the numbered steps in the trajectory below.
    }
    \label{fig:real_word}
    \vspace{-15pt}
\end{figure*}

In this section, we conduct extensive experiments to validate our proposed DreamNav in terms of its capability and feasibility. We first describe the experimental details (Sec.~\ref{sec:experimental_details}), then compare DreamNav with state-of-the-art methods in simulation environments (Sec.~\ref{sec:experimental_sim}), and further evaluate its feasibility in real-world environments (Sec.~\ref{sec:experimental_real}). Finally, we perform ablation studies on key components of DreamNav (Sec.~\ref{sec:experimental_ablation}).

\subsection{Experimental Details}\label{sec:experimental_details}
\textbf{Evaluation Benchmark.} In continuous simulated environments, the Val-Unseen split of the R2R-CE~\cite{vlnce} dataset within the Habitat simulator~\cite{habitat} serves as a widely recognized benchmark for evaluation. Following the experimental protocol of prior work~\cite{navgpt, mapgpt, mc_gpt,opennav,smartway}, we evaluate on 613 trajectories across 11 unseen environments. Besides, to evaluate real-world performance, we followed~\cite{opennav,navid} and designed a comprehensive benchmark comprising four indoor scenes: Office, Corridor, Classroom, and Auditorium. For each environment, we designed five tasks with varying instructions and initial positions, and each task was executed three times.

\begin{table}[t] 
\centering \caption{\footnotesize Comparison with state-of-the-art methods on the Val-Unseen split of R2R-CE. -- indicates that the original paper did not report the corresponding metric. \textbf{Bold} denotes the best performance across all methods, while \underline{Underlined} indicates the second-best. The same notation applies to all subsequent tables.} 
\label{tab:r2r-ce} 
\resizebox{\columnwidth}{!}{
\begin{tabular}{lcccccccc} 
\toprule
\rowcolor{mygray}
\multirow{2}{*}{\textbf{}} & \multicolumn{3}{c}{\textbf{Observation}} & \multicolumn{5}{c}{\textbf{VLN-CE R2R Val-Unseen}} \\ 
\rowcolor{mygray}
& \textbf{Pano.} & \textbf{Egoc.} & \textbf{Odo.} & \textbf{TL} & \textbf{NE$\downarrow$} & \textbf{OSR$\uparrow$} & \textbf{SR$\uparrow$} & \textbf{SPL$\uparrow$} \\

\midrule
\multicolumn{9}{c}{\textbf{Zero-Shot Pano}} \\ 
\midrule

NavGPT-CE$^*$\cite{navgpt}     & $\checkmark$ &  &  & --    & 8.37 & 26.90  & 16.30  & 10.20  \\
DiscussNav-CE$^*$\cite{discuss} & $\checkmark$ &  &  & 6.27 & 7.77 & 15.00  & 11.00  & 10.51 \\
MapGPT-CE$^*$\cite{mapgpt}     & $\checkmark$ &  & $\checkmark$ & 12.63 & 8.16 & 21.00  & 7.00   & 5.04  \\
InstructNav\cite{instructnav}       & $\checkmark$ &  &  & 7.74 & \underline{6.89} & --     & \underline{31.00}  & \underline{24.00}  \\
Open-Nav$^*$\cite{opennav}     & $\checkmark$ &  &  & 7.68 & \textbf{6.70} & 23.00  & 19.00  & 16.10  \\
SmartWay$^*$\cite{smartway}      & $\checkmark$ &  &  & 13.09 & 7.01 & \textbf{51.00}  & 29.0  & 22.46 \\

\midrule
\multicolumn{9}{c}{\textbf{Zero-Shot Egoc}} \\  
\midrule

Open-Nav$^{*{\dagger}}$     & & $\checkmark$ &  & -- & 9.62 & 13.00  & 7.00  & 5.77  \\
InstructNav$^{\dagger}$       & & $\checkmark$ &  & -- &  9.20 & 47.00     & 17.00  &  11.00  \\
CA-Nav\cite{ca_nav}            &  & $\checkmark$ & $\checkmark$ & --    & 7.58 & \underline{48.00}  & 25.30  & 10.80  \\
\textbf{Ours} &  & \textbf{$\checkmark$} &  & 8.20 & 7.06 & 40.95 & \textbf{32.79} & \textbf{28.95} \\
\bottomrule 

\end{tabular}} 
\par\vspace{0.5ex}
{\raggedright\footnotesize
$^*$ denotes methods using the point-level predictor.\\
$^{\dagger}$ self‑implemented results with panoramic inputs replaced by egocentric.\par}
\vspace{-3mm}
\end{table}

\textbf{Evaluation Metrics.} In simulated environments, the performance of VLN tasks is evaluated using the following metrics: Trajectory Length~(TL), the total length of the executed path; Navigation Error (NE), the shortest-path distance between the final position and the goal; Oracle Success Rate~(OSR), the proportion of episodes in which any point along the trajectory falls within the 3 meters success threshold; Success Rate~(SR), the proportion of episodes where the agent stops within 3 meters of the goal; Success-weighted Path Length~(SPL), a path-efficiency metric weighting success by the ratio of the shortest-path distance to the actual path length. In real-world environments, we adopt SR as the primary metric, defining success as stopping within 2 meters of the goal.

\textbf{Implementation Details.} The EgoView Corrector employs a Micro-Adjust Controller that performs 30° rotations with up to two turns under a threshold of $\theta=0.1$, and a Macro-Adjust Expert that performs 90° rotations with up to three turns. The Trajectory Predictor is configured with a Candidate Trajectory Number of 4, and the Imagination Predictor with an Imagination Rollout Length of 18 and an image resolution of 320$\times$448. The Macro-Adjust Expert, Navigator, and Execution Expert are implemented with {GPT-4o}~\cite{gpt_4o}, and the Narration Expert with {Qwen-VL-Max-Latest}~\cite{qwen}. In the Habitat simulator, the horizontal field of view (HFOV) is 69°, the agent height is 1.25 m, the image resolution is 480×640, and the camera is tilted downward by 15°. For real‑world evaluation, the system is deployed on an AgileX LIMO platform. All experiments are conducted on a single NVIDIA RTX 4090 GPU.

\begin{table}[t]
\centering
\caption{\footnotesize Comparison with baseline methods in real-world environments. Results are reported as success counts over five trials per environment. The last column summarizes performance across all 20 trials.}
\label{tab:realworld}
\footnotesize
\resizebox{\columnwidth}{!}{
\begin{tabular}{lccccc}
\toprule
\rowcolor{mygray}
\textbf{} & \textbf{Office} & \textbf{Corridor} & \textbf{Classroom} & \textbf{Auditorium} & \textbf{All}\\
\midrule
Navid\cite{navid}     & 0/5 & 0/5 & \underline{2/5} & 1/5 & 3/20 \\
Open-Nav\cite{opennav}  & \underline{1/5} & \underline{2/5} & 1/5 & \underline{2/5} & \underline{6/20} \\
\textbf{Ours}      & \textbf{2/5} & \textbf{4/5} & \textbf{3/5}  & \textbf{3/5} & \textbf{12/20} \\
\bottomrule
\end{tabular}}
\vspace{-3mm}
\end{table}

\subsection{Evaluation in Simulated Environments}\label{sec:experimental_sim}

 In Table~\ref{tab:r2r-ce}, we benchmark DreamNav against state-of-the-art zero-shot VLN methods using panoramic and egocentric views, respectively. We need to note that the OSR metrics are directly associated with TL, where longer trajectories enlarge the search space and increase overshooting. 
 Therefore, the subsequent discussion focuses on the SR and SPL metrics.

 Our results highlight the advantages of egocentric over panoramic observations as they offer lower sensing costs and stronger performance. Notably, \textbf{DreamNav surpasses all panoramic‑based methods;} compared with the strongest panoramic competitor, InstructNav~\cite{instructnav} (SR: 31.00\%, SPL: 24.00\%), it achieves improvements of 1.79\% in SR and 4.95\% in SPL. Besides,  {human‑like capabilities, specifically trajectory‑level strategy and active imagination, further enhance zero‑shot VLN performance}, as evidenced by \textbf{DreamNav’s superiority over all egocentric‑based baselines}. For instance, the strongest egocentric baseline, CA‑Nav~\cite{ca_nav}, which still relies on odometry information, lags behind DreamNav by 7.49\% in SR and 18.15\% in SPL. The performance gains of DreamNav are not attributable to input modality. In contrast, directly adapting panoramic‑based VLN models to operate under a purely egocentric input setting leads to substantial degradation: when only using the egocentric view, the SR and SPL of Open-Nav~\cite{opennav} drop by 12.00\% and 10.33\%, respectively. 
 Nevertheless, unlike panoramic methods, the absence of global context in egocentric inputs hinders reorientation once the target is lost, inducing cumulative drift and higher NE in DreamNav (7.06m vs. InstructNav: 6.89m; Open‑Nav: 6.70m; SmartWay: 7.01m), pointing to an avenue for refinement.

\subsection{Evaluation in Real-world Environments}\label{sec:experimental_real} 
In real-world implementation, we compare DreamNav with Open-Nav~\cite{opennav}, the latest open-source zero-shot VLN method, and Navid~\cite{navid}, a representative supervised egocentric baseline. As shown in Table~\ref{tab:realworld}, \textbf{zero-shot VLN systems demonstrate high robustness and effectiveness}. Notably, zero-shot variants, including DreamNav and Open-Nav, achieve stable performance across diverse settings, whereas the supervised baseline Navid exhibits pronounced sensitivity to novel environments, with success rates dropping to 0\% in office and corridor scenarios. Furthermore, \textbf{DreamNav demonstrates higher reliability}, with overall success rates exceeding Open‑Nav by 30\% and surpassing the supervised baseline Navid by 45\%. We visualize representative task examples across four environments in Fig.~\hyperref[fig:real_word]{4 (a–d)}, with instruction difficulty increasing progressively.

\begin{table}[t]
\centering
\caption{\footnotesize Effect of EgoView Corrector. MAE denotes the Macro-Adjust Expert and MAC denotes the Micro-Adjust Controller.}
\label{tab:EgoView}
\footnotesize
\begin{tabular}{lcccccc}
\toprule
\rowcolor{mygray}
\multirow{2}{*}{\textbf{}} & \multicolumn{5}{c}{\textbf{VLN-CE R2R Val-Unseen}} \\ 
\rowcolor{mygray}
& \textbf{TL} & \textbf{NE$\downarrow$} & \textbf{OSR$\uparrow$} & \textbf{SR$\uparrow$} & \textbf{SPL$\uparrow$} \\
\midrule
Ours (w/o MAE+MAC)   &  7.60   &  9.87   &  10.00   &  6.00   &  5.43 \\
Ours (only MAC)      &  8.07   &  8.78   &  14.00   &  12.00  &  9.49   \\
Ours (only MAE)      &  8.26   &  \underline{8.05}   & \underline{22.00}   &  \underline{16.00}  &  \underline{14.17}   \\
Ours (with MAE+MAC)  &  8.29 &  \textbf{6.44} &  \textbf{45.00}   &  \textbf{35.00}  & \textbf{30.05} \\

\bottomrule
\end{tabular}
\vspace{-3mm}
\end{table}

\subsection{Ablation Study} \label{sec:experimental_ablation}
We conduct ablations on 100 randomly selected evaluation episodes to analyze the effect of each module in DreamNav.

\textbf{Effect of the EgoView Corrector.} Our EgoView Corrector serves as a key component for enabling effective zero‑shot VLN with purely egocentric observations. We evaluate the EgoView Corrector through four configurations: (i) removing it entirely, (ii) retaining only the Macro‑Adjust Expert (MAE), (iii) retaining only the Micro-Adjust Controller (MAC), and (iv) using the full module. As shown in Tab.~\ref{tab:EgoView}, both the MAE and the MAC offer clear benefits: setting (ii) that only contains the MAE improves SR by 10\% and SPL by 8.74\%, whereas setting (iii) introducing only the MAC increases SR by 6\% and SPL by 4.06\% compared with setting (i) without the entire EgoView Corrector. Moreover, the complementary roles of the MAE and the MAC further facilitate effective egocentric navigation. Notably, the complete setting (iv) (SR:~35\%; SPL:~30.05\%) outperforms all other variants by at least 19\% in SR and 15.88\% in SPL.

\textbf{Effect of Candidate Trajectory Number.}
In our Trajectory Predictor, CTN balances the quality of candidate trajectories against computational efficiency. To examine the impact of CTN, we conduct an ablation study by varying its value from 1 to 6 while keeping all other factors fixed. As shown in Figure 5(a), setting the CTN to 4 represents the optimal trade‑off between computational cost and navigation performance. Specifically, the performance improvement is most significant when the CTN is increased from 1 to 4, but increments beyond 4 only bring negligible improvements. This diminishing return can be attributed to the structural properties of indoor environments, where the egocentric view rarely presents more than four distinct navigable branches.



 \begin{figure}[t]
   \centering
   \includegraphics[width=1\linewidth]{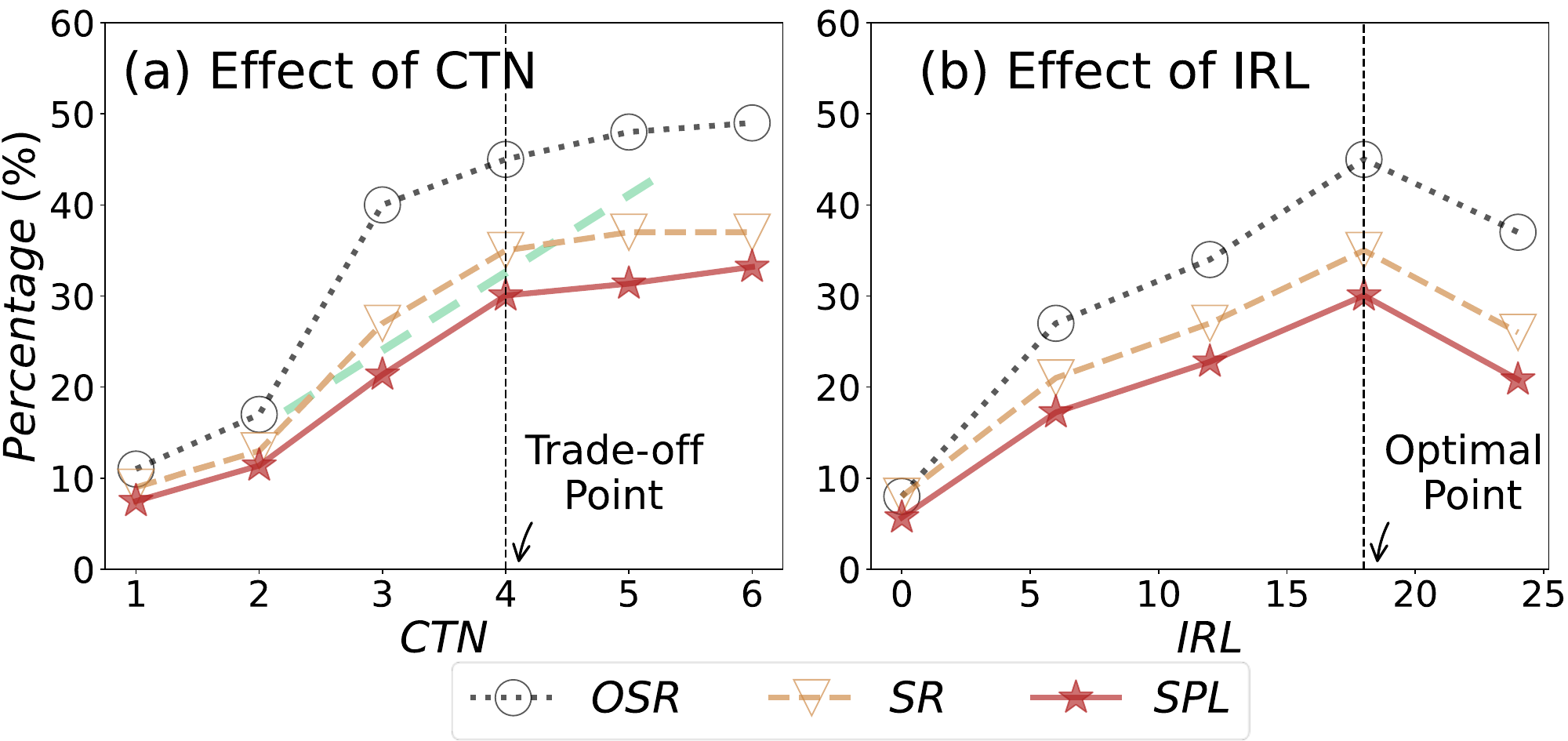}
   \vspace{-15pt}
      \caption{\footnotesize Ablation analysis of two key hyperparameters in DreamNav: (a) Candidate Trajectory Number and (b) Imagination Rollout Length.}
    \label{fig:ablation}
    \vspace{-18.5pt}
\end{figure}
\textbf{Effect of Imagination Rollout Length.} IRL is an important hyperparameter in our Imagination Predictor, balancing beneficial long‑range foresight against detrimental accumulated uncertainty. To investigate the impact of IRL, we conduct an ablation study with all other settings fixed, varying its value over {0, 6, 12, 18, 24}, where 0 serves as the no‑imagination baseline. As shown in Fig.~\hyperref[fig:ablation]{5 (b)}, setting IRL to 18 provides the optimal performance. Specifically, when the IRL is less than 18, longer prospective simulations enhance the agent's decision-making ability by providing richer foresight. However, when the IRL is greater than 18, the agent's performance begins to deteriorate because long periods of imagination accumulate instability and noise in the tail, which blurs meaningful information and ultimately impairs the reliability of decisions.

\section{CONCLUSION}
In this paper, we present DreamNav, the first unified framework that integrates trajectory-level planning with active imagination to investigate the feasibility of achieving intelligent navigation from egocentric observations alone. Within this framework, the EgoView Corrector enables DreamNav to navigate using only lightweight egocentric observations, thus reducing perception overhead. Besides, our Trajectory Predictor addresses the semantic misalignment challenge by generating trajectory-level action policies aligned with instruction semantics to ensure globally coherent navigation. Furthermore, an Imagination Predictor enables long-range reasoning in zero-shot VLN by activating prospective rollouts. Experimental results demonstrate that egocentric observation alone can produce strong performance, with DreamNav surpassing zero‑shot VLN methods using panoramic inputs and achieving substantial improvements over existing egocentric baselines augmented with additional information. We hope our work can inspire more efforts toward simple yet effective VLN frameworks.



\bibliographystyle{IEEEtran}
\bibliography{references}

\end{document}